\documentclass[runningheads]{llncs}
\pdfoutput=1

\usepackage[numbers, sectionbib]{natbib}
\usepackage[utf8]{inputenc} 
\usepackage{hyperref}       
\usepackage{url}            
\usepackage{booktabs}       
\usepackage{amsfonts}       
\usepackage{nicefrac}       
\usepackage{microtype}      
\usepackage{xcolor}         
\usepackage{tikz}
\usetikzlibrary{arrows}
\usepackage{amsmath}
\usepackage{mathtools}
\usepackage{bbm}
\usepackage{amssymb}
\usepackage{algorithm}
\usepackage{algpseudocode}
\usepackage{caption}
\usepackage{subcaption}
\usepackage{graphicx}
\usepackage{float}
\usepackage{placeins}
\usepackage{enumitem}

\graphicspath{{./img/}}

\title{The Feature-First Block Model}

\author{Lawrence Tray \inst{1} \and Ioannis Kontoyiannis \inst{2}}

\institute{Department of Engineering, University of Cambridge, \email{lpt30@cantab.ac.uk} \and 
Statistical Laboratory, University of Cambridge, \email{yiannis@maths.cam.ac.uk}}

\newcommand{\Xcal}{\mathcal{X}}
\newcommand{\Lcal}{\mathcal{L}}
\newcommand{\Bcal}{\mathcal{B}}
\newcommand{\Gcal}{\mathcal{G}}
\newcommand{\Dcal}{\mathcal{D}}
\newcommand{\Fcal}{\mathcal{F}}

\newcommand{\Tcal}{\mathcal{T}}

\newcommand{\one}{\mathbbm{1}}
\newcommand{\Gaussian}{\mathcal{N}}
\newcommand{\indep}{\perp \!\!\! \perp}

\newcommand{\Expect}{\mathbb{E}}
\DeclareMathOperator*{\argmax}{arg\,max} 

\def\imagebox#1#2{\vtop to #1{\null\hbox{#2}\vfill}}

\begin{document}
	
	\maketitle
	
	\begin{abstract}
Labelled networks are an important class of data,
naturally appearing
in numerous applications in science and engineering.
A typical inference goal is to determine how the vertex labels
(or {\em features}) affect the network's structure.
In this work, we introduce a new generative model, the feature-first block model (FFBM),
that facilitates the use of rich queries on labelled networks.
We develop a Bayesian framework and devise a two-level Markov chain Monte 
Carlo approach to efficiently sample from the
relevant posterior distribution of the FFBM parameters. This allows us to infer if and how the observed vertex-features affect macro-structure.
We apply the proposed methods to a variety of network data
to extract the most important features along which the vertices
are partitioned. The main advantages of the proposed approach are
that the whole feature-space is used automatically and that features
can be rank-ordered implicitly according to impact.


\end{abstract}

	\section{Introduction}

Many real-world networks exhibit strong community structure, with most nodes belonging to densely connected clusters. 
Finding ways to recover the latent communities from the observed graph is an important
task in many applications,
including
compression \cite{cluster-compression} and link prediction 
\cite{link-prediction}.
In this work, we examine vertex-labelled networks, 
referring to the labels as {\em features}. A typical goal is to determine whether a given feature impacts graphical structure. Answering this requires a random graph model; the standard is the stochastic block model (SBM) \cite{vanilla-sbm}. Numerous variants of the SBM  have been proposed, e.g., the MMSBM \cite{mixed-membership-sbm} and OSBM \cite{overlapping-sbm}, but these do not include features in the graph generation process.

To analyse a labelled network using one of the simple SBM variants, a typical procedure would be to partition the graph into blocks grouped by distinct values of the feature of interest. The associated model can then be used to test for evidence of heterogeneous connectivity between the feature-grouped blocks. Nevertheless, this approach can only consider disjoint feature sets and the feature-grouped blocks are often an unnatural partition of the graph.

We would instead prefer to partition the graph into its most natural blocks and then find which of the available features -- if any -- best predict the resulting partition. Thus motivated, we present a novel framework for modelling labelled networks, which we call the feature-first block model (FFBM). This is an extension of the SBM to labelled networks.

	\section{Preliminaries}

We first need a model for community-like structure in a network. For this we adopt the widely-used stochastic block model (SBM). This is a latent variable model where each vertex belongs to a single block and the probability two vertices are connected depends only on the block memberships of each.
Specifically, we will use the microcanonical variant of the SBM, proposed by Peixoto \cite{Peixoto-Bayesian-Microcanonical}. To allow for degree-variability between members of the same block, we adopt the following degree-corrected 
formulation (DC-SBM):

\begin{definition}[Microcanonical DC-SBM]
	\label{defn:microcan-dc-sbm}
	Let $N \geq 1$ denote the number of vertices in an undirected graph
with $E$ edges. The block memberships are encoded by a vector $b \in [B]^N$,
where $B$ is the number of non-empty blocks.\footnote{For each integer $K\geq 1$, we use the notation $[K]:=\{1,2,\ldots,K\}$.}
	Let $e=(e_{rs})$ be the $B \times B$ symmetric matrix of edge counts 
between blocks, such that $e_{rs}$ is the number of edges from block $r$ to 
block $s$. 
	Let $k =(k_i)$ denote a vector of length $N$, with $k_i$ being the degree of vertex $i$.

The graph's adjacency matrix $A \in \{0,1\}^{N \times N}$ is generated 
by placing edges uniformly at random, conditional
on the constraints imposed by $b$, $e$ and $k$ being satisfied.
Specifically, if $A \sim \mbox{\rm DC-SBM}_{\rm MC} (b, e, k)$,
then with probability~1 it satisfies,
for all $r,s\in[B]$
and all $i\in[N]$:
	\begin{equation}
		e_{rs} = \sum_{i, j \in [N]} A_{ij} 
	\one \{b_i = r\} \one \{b_j = s\} 
		\qquad 
		\textrm{and} \qquad
		k_i = \sum_{j \in [N]} A_{ij}.
		\label{eqn:sbm-constraints}
	\end{equation}
\end{definition}

	\section{Feature-First Block Model}

\begin{figure}[!ht]
	\centering
	\begin{tikzpicture}[
		roundnode/.style={circle, draw=black, minimum size=7mm},
		squarednode/.style={rectangle, draw=black, minimum size=7mm}
		]
		\node[roundnode] (X) at (0, 0) {$X$};
		\node[squarednode] (b) at (3, 0) {$b$};
		\node[roundnode] (A) at (6, 0) {$A$};
		
		\draw[->] (X.east) -- node[above] {$\theta$} (b.west);
		\draw[->] (b.east) -- node[above] {$\psi$}(A.west);
	\end{tikzpicture}
	\caption{The Feature-First Block Model (FFBM)}
	\label{fig:ffbm}
\end{figure}
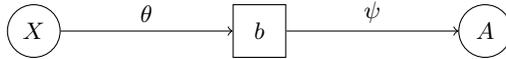

In this section we propose a novel generative model for labelled networks. We call this the feature-first block model (FFBM),
illustrated in Figure~\ref{fig:ffbm}.

Let $N$ denote the number of vertices, $B$ the number of blocks
and ${\cal X}$ the set of values each feature can take.
We define the vector $x_i \in \Xcal^D$ as the feature vector for vertex $i$, 
where $D$ is the number of features associated with each vertex.
For example, in the datasets we analyse, we deal with binary feature flags
(denoting the presence/absence of each feature),
so $\Xcal = \{0, 1\}$. We write $X$ for the $N\times D$ {\em feature matrix} containing
the feature vectors $\{x_i\}_{i=1}^{N}$ 
as its rows.

For the FFBM, we start with the feature matrix $X$ and generate a random
vector of block memberships $b \in [B]^N$. For each vertex $i$, the
block membership $b_i\in[B]$ is generated based on the feature
vector $x_i$, independently between vertices. The conditional
distribution of $b_i$ given $x_i$ also depends on a collection
of weight vectors $\theta=\{w_k\}_{k=1}^B$, where each
$w_k$ has dimension $D$. We will later find it convenient
to write $\theta$ as a $B \times D$ matrix of weights $W$. Specifically, 
the distribution of $b$ given $X$ and $\theta$ is,
\begin{equation}
	p(b| X, \theta) = \prod_{i \in [N]} p(b_i | x_i, \theta) = \prod_{i \in [N]} \phi_{b_i} (x_i; \theta)
	= \prod_{i \in [N]} \frac{\exp\left(w_{b_i}^T x_i\right)}{\sum_{k \in [B]} \exp \left( w_k^T x_i\right)}.
\end{equation}
Note that $\phi_{b_i}$ has the form of a softmax activation function.
More complex models based on different choices for the distributions
$\phi_{b_i}$ above are also possible, but then deriving meaning from 
the inferred parameter distributions is more difficult. 

Once the block memberships $b$ have been generated, we then draw the 
graph $A$ from the microcanonical DC-SBM with additional parameters 
$\psi = \{\psi_e, \psi_k\}$:
\begin{equation}
	A \sim \textrm{DC-SBM}_{\textrm{MC}} (b, \psi_e, \psi_k).
	\label{eqn:A-generation}
\end{equation}

\subsection{Prior selection}

To complete the description of our Bayesian framework,
priors on $\theta$ and $\psi$ must also be specified. 
We place a Gaussian prior on $\theta$ such that
each element of $\theta$ has an independent ${\cal N}(0,\sigma_\theta^2)$
prior, with hyperparameter $\sigma_\theta^2$:
\begin{equation}
	p(\theta) \sim \Gaussian \left( \theta ; 0, \sigma_\theta^2 I \right).
	\label{eqn:theta-prior}
\end{equation}
This choice of prior gives a very
simple form for the conditional distribution of the block membership vector $b$ given $X$; it is a uniform distribution:
\begin{equation}
	p(b | X) = \int p(b | X, \theta) p(\theta) d\theta = B^{-N}.
	\label{eqn:b-pseudo-prior}
\end{equation}
The proof is given in Appendix~\ref{appdx:b|x}. This is an important simplification as evaluating $p(b | X)$ does not require an expensive  integration over $\theta$ nor does it depend on $X$.
Peixoto \cite{Peixoto-Bayesian-Microcanonical} proposes careful choices for 
the priors on the additional microcanonical SBM parameters $\psi$, which we adopt without repeating their exact form here. 
The idea is to write the joint distribution on $(b, e, k)$ as a product of 
conditionals, $p(b, e, k) = p(b) p(e | b) p(k | e, b)= p(b) p(\psi | b)$. 
In our case, conditioning on $X$ is also necessary, leading to,
$
	p(b, \psi | X) = p(b | X) p(\psi | b, X) = p(b | X) p(\psi | b),
$
where we used the fact $\psi$ and $X$ are conditionally 
independent given $b$.
All that concerns the main argument is that $p(\psi|b)$ has
an easily computable form.

	\section{Inference}
\label{sec:inference}

Having completed the definition of the FFBM, we wish to leverage it 
to perform inference. Specifically, given a labelled network $(A, X)$, we wish to infer if and how the observed features $X$ impact the graphical structure $A$. Formally,
this means characterising the posterior distribution:
$
p(\theta|A, X) \propto p(\theta) \cdot p(A | X, \theta).
$
Although the prior is easily computable, 
computing the likelihood 
requires summing over all latent block-states, 
$p(A| X, \theta) = \sum_{b \in [B]^N} p(A | b) P(b | X, \theta)$, which is 
clearly impractical. In fact, this
approach is doubly intractable as we would also 
need to compute the normalising constant $p(A|X)$.
Therefore, following standard Bayesian practice,
instead we aim to draw samples from the posterior,
\begin{equation}
	\label{eqn:theta-target}
	\theta^{(t)} \sim p(\theta | A, X).
\end{equation}
We propose an iterative Markov chain Monte Carlo
(MCMC) approach to obtain these samples
$\{\theta^{(t)}\}$. We first draw a sample $b^{(t)}$ 
from the block membership posterior,
and then use $b^{(t)}$ to obtain a corresponding
sample $\theta^{(t)}$:
\begin{equation}
	b^{(t)} \stackrel{\rm distr}{\approx} p \big( b | A, X \big) 
	\quad \textrm{then} \quad
	\theta^{(t)} \stackrel{\rm distr}{\approx} 
	p\big(\theta | X, b^{(t)} \big),
\end{equation}
where these approximations become exact as
the number of MCMC iterations $t\to\infty$.
As described in the following subsections,
this can be implemented through a two-level
Markov chain via the Metropolis-Hastings (MH) 
algorithm \cite{hastings-alg}.
The splitting of the Markov chain into two levels allows us to side-step the summation over
all latent $b \in [B]^N$ required to directly compute the likelihood, $p(A| X, \theta)$.
The resulting $\theta^{(t)}$ samples are asymptotically
unbiased in that the expectation of 
their distribution converges to the true posterior:
\begin{equation}
\lim_{t\to\infty}
\Expect_{b^{(t)}} \left[p \left( \theta | X, b^{(t)} \right) \right] = \sum_{b \in [B]^N} p(\theta | X, b) p(b | A, X) = p(\theta | A, X).
\label{eqn:theta-unbiased}
\end{equation}
This is an example of a pseudo-marginal approach;
see, e.g., Andrieu and Roberts~\cite{pseudo-marginal} 
for a detailed rigorous derivation based on~(\ref{eqn:theta-unbiased}).
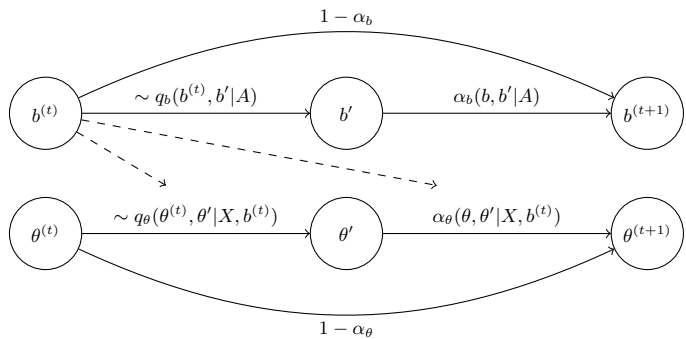
\begin{figure}[!ht]
	\centering

	\begin{tikzpicture}[
		scale=0.8, every node/.style={transform shape},
		roundnode/.style={circle, draw=black, minimum size=12mm},
		squarednode/.style={rectangle, draw=black, minimum size=12mm}
		]
		\node[roundnode] (b0) at (0, 2) {$b^{(t)}$};
		\node[roundnode] (b1) at (5, 2) {$b'$};
		\node[roundnode] (b2) at (10, 2) {$b^{(t+1)}$};
		\node[roundnode] (t0) at (0, 0) {$\theta^{(t)}$};
		\node[roundnode] (t1) at (5, 0) {$\theta'$};
		\node[roundnode] (t2) at (10, 0) {$\theta^{(t+1)}$};
		
		\draw[->] (b0) to node[above] {$\sim q_b ( b^{(t)}, b' | A )$} (b1);
		\draw[->] (b1) to node[above] {$\alpha_b (b, b' | A )$} (b2);
		\draw[->] (b0) [out=25, in=155] to node[above] {$1-\alpha_b$} (b2);
		
		\draw[->] (t0) to node[above] {$\sim q_\theta(\theta^{(t)}, \theta' | X, b^{(t)})$} (t1);
		\draw[->] (t1) to node[above] {$\alpha_\theta (\theta, \theta' | X, b^{(t)})$} (t2);
		\draw[->] (t0) [out=-25, in=-155] to node[below] {$1-\alpha_\theta$} (t2);
		
		\draw[dashed, ->] (b0) to (2, 0.8);
		\draw[dashed, ->] (b0) to (6.5, 0.8);
		
	\end{tikzpicture}
	\caption{$\theta$-sample generation.}
	\label{fig:samp-sequence}
\end{figure}
 
Figure \ref{fig:samp-sequence} shows an overview of the proposed method, with $q$ and $\alpha$ denoting the MH proposal distribution and acceptance probability respectively.
Note the importance of the simplification in~(\ref{eqn:b-pseudo-prior}). 
As evaluating $p(b| X)$ does not depend on $X$, 
we do not need $X$ to sample $b$.
And on the other level, in order to obtain 
samples for $\theta$
we use only $b$ but not $A$, as $(\theta \indep A )| b$. 

\FloatBarrier
\subsection{Sampling block memberships}

To generate the required $b$-samples, we adopt the MCMC
procedure of
\cite{Peixoto-MCMC},
which relies on writing the posterior in the following form,
\begin{equation}
	p(b | A, X) \propto p(A | b, X) \cdot p(b | X) = \pi_b(b),
\end{equation}
where $\pi_b(\cdot)$ denotes the un-normalised target density.
Since we are using the microcanonical SBM formulation, there is only one 
value of $\psi$ that is compatible with the given $(A, b)$ pair;
recall the constraints in~(\ref{eqn:sbm-constraints}).
We denote this value $\psi^* = \{\psi_k^*, \psi_e^*\}$. Therefore, 
the summation over all $\psi$ needed to evaluate $p(A | b, X)$ reduces to just the single $\psi^*$ term:
$p(A | b, X) = \sum_{\psi} \nolimits p(A , \psi | b, X) = p(A, \psi^* | b, X)$.
In
this context, the microcanonical entropy of the configuration $b$
is,
\begin{equation}
	S(b) \coloneqq - \log \pi_b(b) = - \Big( \log p(A | b, \psi^*) + \log p(\psi^*, b | X) \Big),
	\label{eqn:dl-form}
\end{equation}
which can be thought of as the optimal
``description length'' of the graph. 
This expression will later be employed 
to help evaluate experimental results. 
The exact form of the proposal $q_b$ is explored thoroughly in
\cite{Peixoto-MCMC} and not repeated here. We use the \verb*|graph-tool| \cite{peixoto_graph-tool_2014}
library for Python, which implements this algorithm.
The only modification is in 
the prior $p(b)$ that we replace with $p(b|X)=B^{-N}$, 
which cancels out in the MH accept-reject step as it is independent of $b$.

\subsection{Sampling feature-to-block generator parameters}
\label{s:sfb}

The target distribution for the required $\theta$-samples 
is the posterior of $\theta$ given the values of the pair $(X, b)$. 
We write this as,
\begin{equation}
	\pi_\theta(\theta) \propto p(\theta | X, b) \propto p(b | X, \theta) p(\theta) \propto  \exp \left( - U(\theta) \right),
	\label{eq:U}
\end{equation}
where $U(\theta)$ denotes the negative log-posterior. Let $y_{ij} \coloneqq \one \left\{ b_i = j \right\}$ and $a_{ij} \coloneqq \phi_j(x_i; \theta)$. 
Discarding constant terms, $U(\theta)$ can be expressed as,
\begin{equation}
	U(\theta) = \left( \sum_{i \in [N]} \sum_{j \in [B]} y_{ij} \log \frac{1}{a_{ij}} \right)
	+ \frac{1}{2\sigma_\theta^2} \|\theta\|^2 = N \cdot \Lcal(\theta) + \frac{1}{2\sigma_\theta^2} \|\theta\|^2;
	\label{eqn:U-form}
\end{equation}
see Appendix \ref{appdx:form-U}. The function $U(\theta)$ is a typical objective function for neural network training. The first term $N \cdot \Lcal(\theta)$ is introduced by the likelihood and represents the cross-entropy between the graph-predicted and feature-predicted block memberships. 
The second term, introduced by the prior, brings a form of regularisation, guarding against over-fitting. In order to draw samples from the posterior 
$\pi_\theta \propto \exp(-U)$ we adopt the Metropolis-adjusted Langevin 
algorithm (MALA) \cite{mala-tweedie}, which uses $\nabla U$ to bias the 
proposal towards regions of higher density. Given the current 
sample $\theta$, a proposed 
new sample $\theta'$ is generated from,
\begin{equation*}
	\theta' \sim q_\theta\big(\theta, \theta'\big) 
	= \Gaussian \big( \theta' ; \theta - h \nabla U(\theta), 2h I \big),
\end{equation*}
where $\xi \sim \Gaussian(0, I)$ and $h$ is a step-size parameter 
which may vary with the sample index.
Without the injected noise term $\xi$, MALA is equivalent to gradient descent. We require $\xi$ to fully explore the parameter space. 
The term $\nabla U$ has an easy to compute analytic form (derived in Appendix \ref{appdx:form-U}).

\subsection{Sampling sequence}
\label{s:ss}

So far, each $\theta^{(t)}$ update has used its corresponding $b^{(t)}$ sample. This means the evaluation of $U^{(t)}$ and $\nabla U^{(t)}$ has high variance, leading to longer burn-in and possibly slower MCMC convergence. The only link between $b^{(t)}$ and $\theta^{(t)}$ is in the evaluation of $U^{(t)}$ and $\nabla U^{(t)}$ which depends only on the matrix $y^{(t)}$ with entries $y_{ij}^{(t)} \coloneqq \one\{b_i^{(t)} = j\}$. We would rather deal with the expectation of each $y_{ij}^{(t)}$:
\begin{equation}
	\Expect \left[ y_{ij}^{(t)} \right] = \Expect_{b^{(t)}} \left[ \one \left( b_{i}^{(t)} = j \right) \right]
	= p(b_i = j | A, X).
\end{equation}
An unbiased estimate for this can be obtained using 
the thinned $b$-samples after burn-in.
Let $\Tcal_b$  denote the retained set of indices 
for the $b$-samples and $\Tcal_\theta$ similarly for the $\theta$-chain. 
The unbiased estimate for $y_{ij}^{(t)}$ is then:
\begin{equation}
	\hat{y}_{ij} \coloneqq \frac{1}{|\Tcal_b|} \sum_{t \in \Tcal_b} y_{ij}^{(t)} = \frac{1}{|\Tcal_b|} \sum_{t \in \Tcal_b} \one\{b_i^{(t)} = j\}.
	\label{eqn:y-hat}
\end{equation}
The same matrix $\hat{y}$ is used in each $\theta^{(t)}$ update step.
This way, it is not necessary to run the $b$ and $\theta$ Markov chains 
concurrently. Instead, we run the $b$-chain to completion and use it 
to generate $\hat{y}$ also allowing us to vary the lengths of each.

\subsection{Dimensionality reduction}
\label{sec:dim-reduction}

The complexity of evaluating $U$ and $\nabla U$ is linear in 
the dimension of the feature space $D$,
so there is computational incentive to reduce $D$.
Given the samples $\left\{ \theta^{(t)} \right\}$, we can compute the empirical mean and standard deviation of each component of $\theta$. 
Switching to the matrix notation $W$ for $\theta$,
let:
\begin{equation}
	\hat{\mu}_{ij} \coloneqq \frac{1}{|\Tcal_\theta|} \sum_{t \in \Tcal_\theta} W_{ij}^{(t)} \qquad \textrm{and} \qquad
	\hat{\sigma}_{ij}^2 \coloneqq \frac{1}{|\Tcal_\theta|} \sum_{t \in \Tcal_\theta} \left( W_{ij}^{(t)} - \hat{\mu}_{ij} \right)^2.
\end{equation}
A simple heuristic to discard the least important features requires specifying a cutoff $c > 0$ and a multiplier $k > 0$. We define the function $\Fcal_i(j)$ 
as in~(\ref{eqn:fij}) and only keep features with indices $d \in \Dcal'$, where $\Dcal'$ is given in~(\ref{eqn:kept-feature-set}).
\begin{align}
	\Fcal_i(j) &\coloneqq (\hat{\mu}_{ij} - k \hat{\sigma}_{ij}, \hat{\mu}_{ij} + k \hat{\sigma}_{ij}) \cap (-c, +c),
	\label{eqn:fij} \\
	\Dcal' &\coloneqq \left\{ j \in [D] : \exists i \in [B] \textrm{ s.t. }  \Fcal_i(j) = \emptyset \right\}.
	\label{eqn:kept-feature-set}
\end{align}
Intuitively, this means discarding any feature $j$ for which 
$(\hat{\mu}_{ij} - k\hat{\sigma}_{ij}, \hat{\mu}_{ij} + k \hat{\sigma}_{ij})$ overlaps with
$(-c, c)$ for all $i$. If we were to use the Laplace approximation for the posterior $p(W_{ij} | A, X) \approx \Gaussian(W_{ij}; \hat{\mu}_{ij}, \hat{\sigma}_{ij}^2)$, then this would be analogous to a hypothesis test on the magnitude of $W_{ij}$ compared to $c$ with multiplier $k$ in~(\ref{eqn:fij}) determining the degree of significance of the result. Conversely, if we want to fix the number of dimensions in our reduced feature set $|\Dcal'|=D'$, the problem then becomes finding the largest value of $c$ such that $|\Dcal'|=D'$ given $k=k_0$:
\begin{equation}
	c^* = \argmax \{c>0\; : \;|\Dcal'| = D', k=k_0\}.
	\label{eqn:c-star}
\end{equation}

	\section{Experimental results}
\label{sec:experiments}

\begin{figure}[!ht]
	\centering
	\begin{subfigure}[t]{0.28\linewidth}
		\centering
		\includegraphics[width=\linewidth]{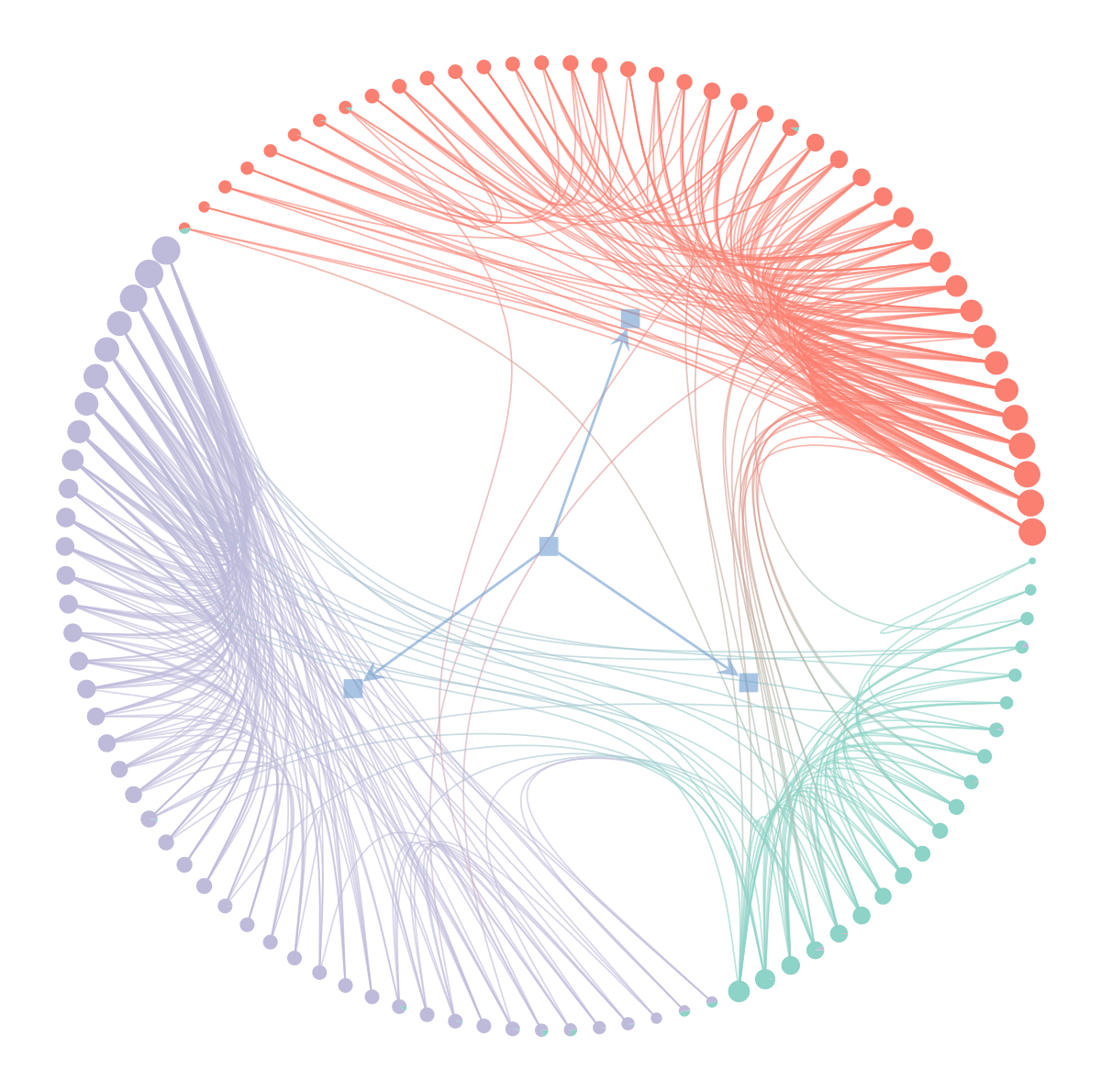}
		\caption{Polbooks}
		\label{fig:polbooks-graph}
	\end{subfigure}
	\hfill
	\begin{subfigure}[t]{0.28\linewidth}
		\centering
		\includegraphics[width=\linewidth]{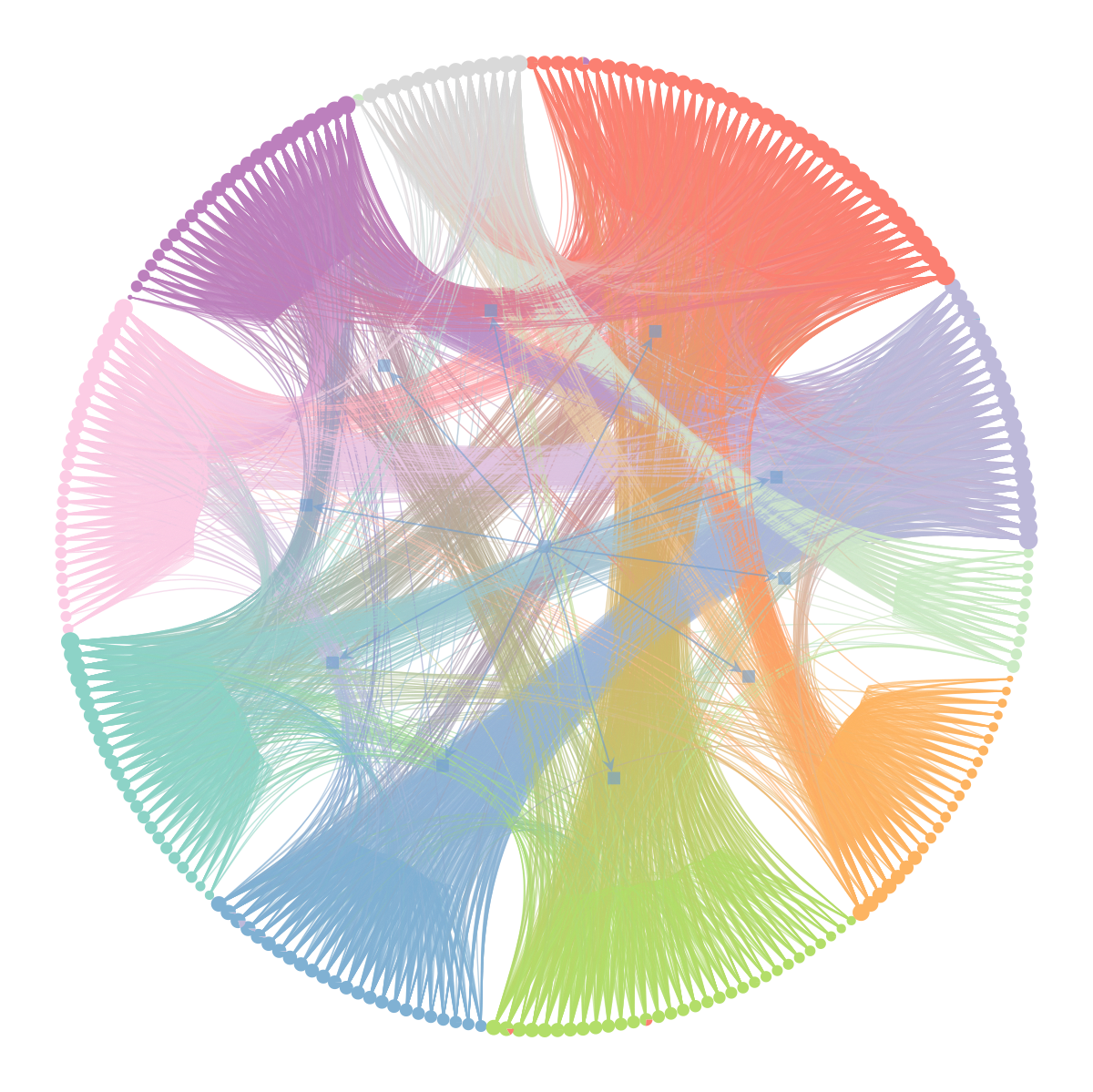}
		\caption{Primary school}
		\label{fig:school-graph}
	\end{subfigure}
	\hfill
	\begin{subfigure}[t]{0.28\linewidth}
		\centering
		\includegraphics[width=\linewidth]{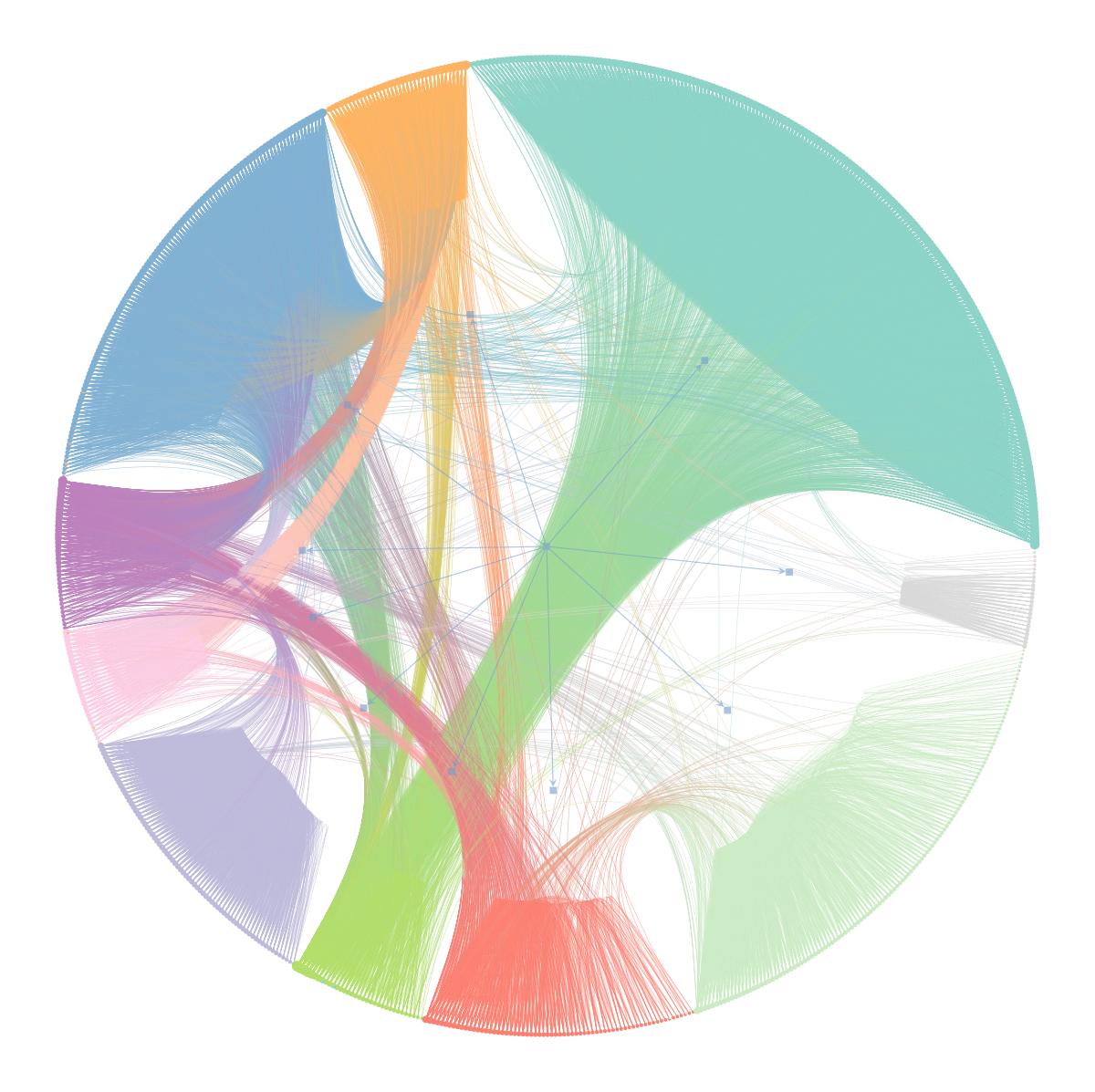}
		\caption{Facebook egonet}
		\label{fig:fb-graph}
	\end{subfigure}
	\begin{subfigure}[t]{0.11\linewidth}
		\centering
		\includegraphics[width=0.8\linewidth]{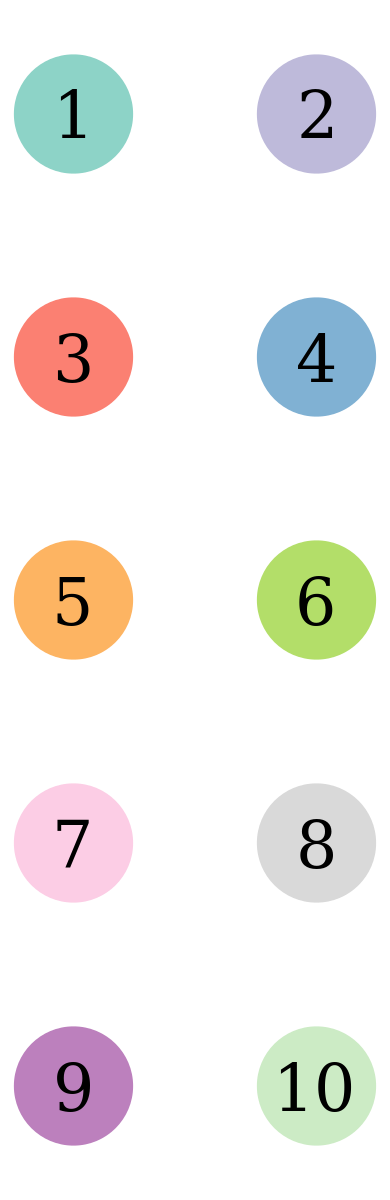}
		\caption{Block Legend}
		\label{fig:10-legend}
	\end{subfigure}
	\caption{Networks laid out and coloured according to inferred block memberships for a given experiment iteration. Visualisation performed using \texttt{graph-tool} \cite{peixoto_graph-tool_2014}.}
	\label{fig:graphs-all}
\end{figure}

We apply our proposed methods to a variety of labelled networks:

\begin{itemize}
	\item \textbf{Political books} \cite{polbooks} ($N=105, E=441, D=3$) -- network of Amazon political books published close to the 2004 presidential election. Two books are connected if they were frequently co-purchased. Vertex features encode the political affiliation of the author (liberal, conservative, or neutral).
	\item \textbf{Primary school dynamic contacts} \cite{schools} ($N=238, E=5539, D=13$) -- network of 238 individuals (students and teachers),
with edges denoting face-to-face contacts at a primary school in Lyon, France. 
The vertex features are class membership (one of 10 values: 1A-5B), gender (male, female), and status (teacher, student). We choose to analyse just the second day of results.
	\item \textbf{Facebook egonet} \cite{fb-snap} ($N=747, E=30025, D=480$) -- network of Facebook users with edges denoting ``friends''.
Vertex features are fully anonymised and encode information about each user's education history, languages spoken, gender, home-town, birthday etc. We focus on the egonet with id 1912.
\end{itemize}
For reference, the inferred partitions for all of these are given on Figure~\ref{fig:graphs-all}.
We employ the following metrics to assess model performance. 
First, the average
description length per entity (nodes and edges) 
$\bar{S}_e$ 
used to gauge the SBM fit is defined as:
\begin{equation}
	\bar{S}_e \coloneqq \frac{1}{(N+E) |\Tcal_b|} \sum_{t\in \Tcal_b} S \left( b^{(t)} \right).
	\label{eqn:mean-dl}
\end{equation}
Next, to assess the performance of the feature-to-block predictor, 
the vertex set $[N]$ 
is partitioned at random so that 
a constant fraction $f$ of vertices form the training set $\Gcal_0$ and 
the remainder form the test set $\Gcal_1$.
The $b$-chain is run using the whole network but only vertices $v \in \Gcal_0$ are 
used for the $\theta$-chain. 
Then the  the average cross-entropy loss 
over each set is used to gauge the quality of the fit,
\begin{equation}
	\bar{\Lcal}_\star \coloneqq \frac{1}{|\Tcal_\theta|} \sum_{t \in \Tcal_\theta} \Lcal_\star^{(t)},
	\quad \textrm{where} \quad
	\Lcal_\star^{(t)} \coloneqq \frac{1}{|\Gcal_\star|} \sum_{i \in \Gcal_\star}\sum_{j \in [B]} \hat{y}_{ij} \log \frac{1}{\phi_j \left(x_i; \theta^{(t)} \right)},
	\label{eqn:cross-entropy-loss}
\end{equation}
where $\star \in \{0, 1\}$ toggles between the training and test sets
and $\hat{y}_{ij}$ is defined in~(\ref{eqn:y-hat}).
Nevertheless, the cross-entropy loss is a coarse measure of fit. 
A new measure, specific to each detected block,
can be defined as follows. Let
$\Bcal_\star(j)$ 
be the set of vertices with maximum a posteriori probability of belonging 
to block~$j$,
$
	\Bcal_\star(j) \coloneqq \{i \in \Gcal_\star : \hat{b}_i = j\},
$
where
$ 
	\hat{b}_i \coloneqq \argmax_j \hat{y}_{ij},
$
and
define the {\em block-accuracy} for block $j$ as,
\begin{equation}
	\eta_\star(j) \coloneqq \frac{1}{|\Bcal_\star (j)| \cdot 
	|\Tcal_\theta| } 
	\sum_{i \in \Bcal_\star (j)}  \sum_{t \in \Tcal_\theta}
	\one \left\{\hat{b}_i = \argmax_j \phi_j \left( x_i; \theta^{(t)} \right) \right\}.
	\label{eqn:accuracy}
\end{equation}
This effectively tests whether the feature-to-block and 
graph-to-block predictions agree in their largest component.
For the higher-dimensional datasets, we also apply the 
dimensionality reduction method 
of Section~\ref{sec:dim-reduction}.  
We then retrain the feature-block predictor using only the retained 
feature set $\Dcal'$, and report the log-loss over the training and 
test sets for the reduced classifier -- 
denoted $\bar{\Lcal}_0'$ and $\bar{\Lcal}_1'$ respectively. 
\begin{table}[!ht]
	\centering
	\caption{Experimental results averaged over $n=10$ iterations (mean $\pm$ std. dev.).}
	\label{tab:results}
	\resizebox{\textwidth}{!}{%
		\begin{tabular}{c|ccc|c|cc|ccc}
			Dataset  & $B$ & $D$ & $D'$ & $\bar{S}_e$ & $\bar{\Lcal}_0$ & $\bar{\Lcal}_1$ & $c^*$ & $\bar{\Lcal}_0'$ & $\bar{\Lcal}_1'$  \\ \hline
			Polbooks & 3 & 3 & -- & $2.250 \pm 0.000$ & $0.563 \pm 0.042$ & $0.595 \pm 0.089$ & -- & -- & -- \\
			School & 10 & 13 & 10 & $1.894 \pm 0.004$ & $0.787 \pm 0.127$ & $0.885 \pm 0.129$ & $1.198 \pm 0.249$ & $0.793 \pm 0.132$ & $0.853 \pm 0.132$ \\
			FB egonet & 10  & 480 & 10 & $1.626 \pm 0.003$ & $1.326 \pm 0.043$ & $1.538 \pm 0.069$ & $0.94 \pm 0.019$ & $1.580 \pm 0.150$ & $1.605 \pm 0.106$
		\end{tabular}
	}
\end{table}
\begin{figure}[!ht]
	\centering
	\begin{subfigure}{0.32\linewidth}
		\centering
		\imagebox{0.9\linewidth}{\includegraphics[width=\linewidth]{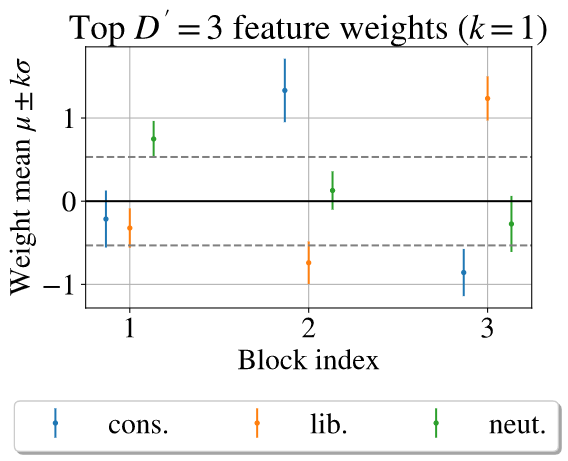}}
		\caption{Political books}
		\label{fig:polbooks-null}
	\end{subfigure}
	\begin{subfigure}{0.32\linewidth}
		\centering
		\imagebox{0.9\linewidth}{\includegraphics[width=\linewidth]{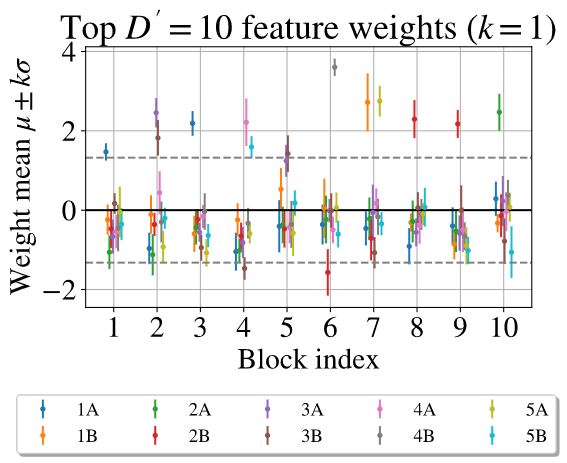}}
		\caption{Primary school}
		\label{fig:school-null}
	\end{subfigure}
	\begin{subfigure}{0.32\linewidth}
		\centering
		\imagebox{0.9\linewidth}{\includegraphics[width=\linewidth]{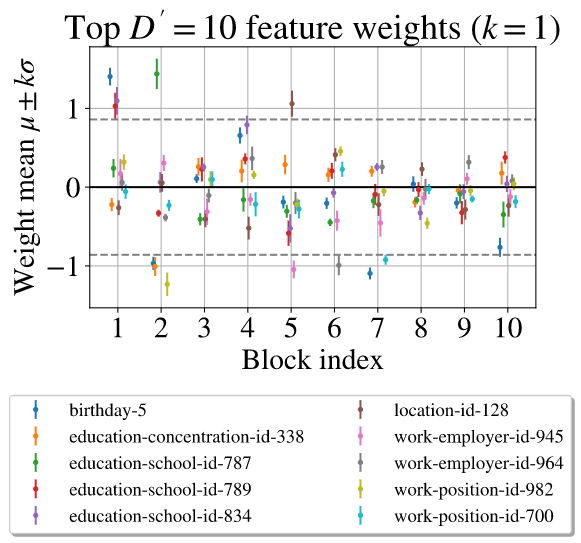}}
		\caption{Facebook egonet}
		\label{fig:fb-null}
	\end{subfigure}
	\caption{Top $D'$ $\theta$-samples for each dataset. Coarse steps on x-axis give block index and the fine steps denote give index. Dotted line is $\pm c^*$.}
\end{figure}
\begin{figure}[!ht]
	\centering
	\begin{subfigure}{0.32\linewidth}
		\centering
		\imagebox{0.8\linewidth}{\includegraphics[width=\linewidth]{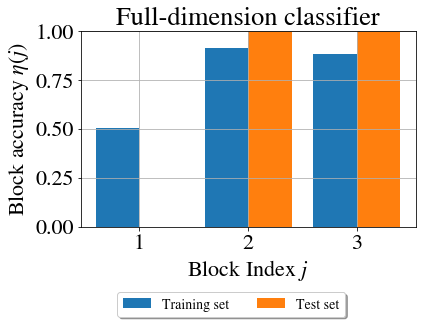}}
		\caption{Political books}
		\label{fig:polbooks-accuracy}
	\end{subfigure}
	\begin{subfigure}{0.32\linewidth}
		\centering
		\imagebox{0.8\linewidth}{\includegraphics[width=\linewidth]{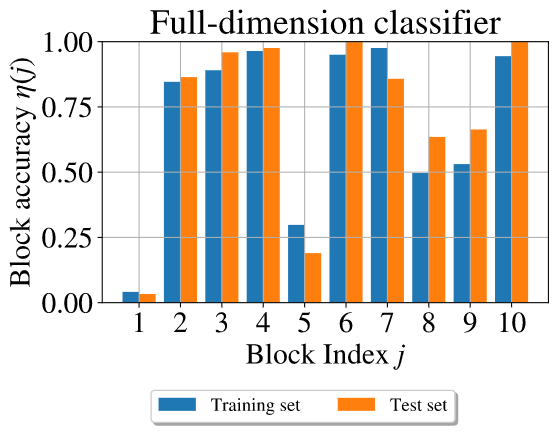}}
		\caption{Primary school}
		\label{fig:school-accuracy}
	\end{subfigure}
	\begin{subfigure}{0.32\linewidth}
		\centering
		\imagebox{0.8\linewidth}{\includegraphics[width=\linewidth]{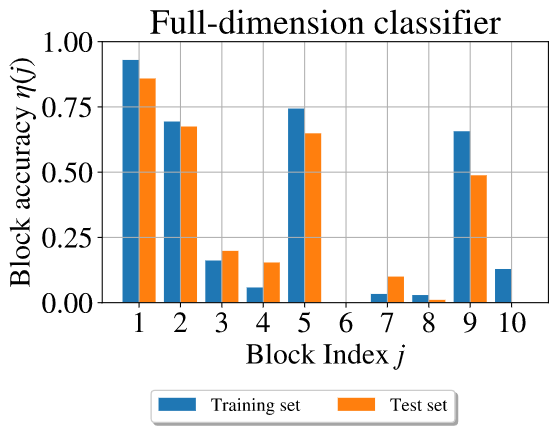}}
		\caption{Facebook egonet}
		\label{fig:fb-accuracy}
	\end{subfigure}
	\caption{Per-block accuracy $\eta(j)$ for each dataset.}
\end{figure}
\FloatBarrier
%
%
%
%

Table~\ref{tab:results} summarises the results for each experiment. 
We see that the dimensionality reduction procedure 
brings the training and test losses closer. This indicates that
the retained features
are indeed well correlated with the underlying graphical 
partition and that the approach generalises correctly. The test loss variance is higher than the training loss variance as the test set is smaller and so more susceptible to variability in its construction.
The average description length per entity of the graph,
$\bar{S}_e$,
has very small variance, suggesting that
the detected communities can be found reliably (to within an arbitrary 
relabelling of blocks).

\paragraph{\textbf{Political books.}}

We choose to partition the network into $B=3$ communities as we only have this many distinct values for political affiliation.
From Figure~\ref{fig:polbooks-null} we see that all 3 blocks have a distinct political affiliation as their largest positive component.  
Furthermore, the training and test losses from Table~\ref{tab:results}  
are very similar and both are low in magnitude. This is strong evidence 
that political affiliation is a very appropriate explanatory 
variable for the overall network structure.
However, from Figure~\ref{fig:polbooks-accuracy} we see that block 1 has low accuracy. 
This suggests that detected block 1 is not solely composed of ``neutral" books but also 
contains some ``liberal" and ``conservative" authors. Examining 
Figure~\ref{fig:polbooks-graph}, we see the majority of paths between blocks 2 and 3 go through block 1.
Block 1 is in effect a bridge between the ``conservative'' and ``liberal'' blocks so it is unsurprising that some of these leak into block 1.

\paragraph{\textbf{Primary school.}}

We choose the number of communities $B=10$, in line with the total number of 
school classes. Only the pupils' class memberships (1A-5B) survive
the dimensionality-reduction process (Figure~\ref{fig:school-null});
gender and teacher/student status have been discarded,
meaning these are poor predictors of overall macro-structure.
Almost all blocks are composed of a single class. 
However, some blocks have two comparably strong classes as their predictors (e.g. blocks 2 and 5). 
Conversely, some classes are found to extend over two 
detected blocks (class 2B spans blocks 8 and 9) but we do 
not have a feature which explains the division.
Figure~\ref{fig:school-accuracy} shows excellent accuracy for most blocks. In fact the only blocks with low accuracy are those with a `school-class'
feature that spans two blocks, such that we cannot reliably distinguish between the two. This is more pronounced when we apply hard classification rather than cross-entropy loss. It is possible that there are unobserved features here
which explain this divide.

\paragraph{\textbf{Facebook egonet.}}

The retained features 
(Figure~\ref{fig:fb-null}) are those that best explain the high-level 
community structure. The majority of these are education related. 
Nevertheless, for $D'=10$ we only have good explanations for some of the detected blocks; several blocks in 
Figure~\ref{fig:fb-null} do not have high-magnitude components. This is further emphasised by the disparate accuracies in Figure~\ref{fig:fb-accuracy}.
For a high-dimensional feature-space, it is likely that a particular
feature may uniquely identify a small set of vertices; if these are all in the same block, then the classifier may overfit despite the penalty imposed by the prior. Indeed, we see in
Figure~\ref{fig:fb-null} that the feature `birthday-5' has a very high weight as it relates to block 1 – but it is unlikely that birthdays determine graphical structure.

	\section{Conclusion}
\label{sec:conclusion}

The proposed Feature-First Block Model (FFBM)
is a new generative model for labelled networks.
It is a hierarchical Bayesian model, 
well-suited for describing how features
affect network structure.
The Bayesian inference tools developed in this 
work facilitate the identification of 
vertex features that are in some way 
correlated with the network's graphical structure.
Consequently, 
finding the features that best describe the most 
pronounced partition, makes it possible in practice
to examine the existence of -- and to make a case for --
causal relationships.

An efficient MCMC algorithm 
is developed for sampling 
from the posterior distribution of
the relevant parameters in the FFBM;
the main idea is to divide up the graph into 
its most natural partition under the associated
parameter values, and then to determine whether 
the vertex features can accurately explain the partition. 
Through several applications on empirical
network data, this approach 
is shown to be effective at extracting and describing 
the most natural communities in a labelled network. 
Nevertheless, it
can only currently explain the structure at the macroscopic
scale. Future work will benefit from extending 
the FFBM to a further hierarchical model,
so that
the structure of the network 
can be explained at all scales of interest.

	\appendix
	\section{Appendix}

\subsection{Derivation of {\boldmath $p(b|X)$}}
\label{appdx:b|x}

We determine the form of $p(b| X)$ by integrating
out the parameters $\theta$. From the definitions, we have:
\begin{align*}
	p(b | X) 
	&= \int p(b , \theta| X) d\theta 
	= \int p(b | X, \theta) p(\theta | X) d\theta
	= \int \prod_{i \in [N] } \phi_{b_i}(x_i; \theta) p(\theta) d\theta \\
	&= \prod_{i \in [N]} \int \frac{\exp(w_{b_i}^T x_i) \prod_{j \in [B]} \Gaussian(w_j; 0, \sigma_\theta^2 I)}{\sum_{k \in [B]} \exp(w_{k}^T x_i)} dw_{1:B}.
\end{align*}
The key observation here is that 
the value of the integral is independent
of the value of $b_i \in [B]$ as
the integrand has the same form regardless of $b_i$. This is
because the prior is the same for each $w_j$. 
Therefore, the integral can only be a function of $x_i$ and $\sigma_\theta^2$,
which means that, as a function of $b$, $p(b|X)\propto 1$. As
$b$ takes values in $[B]^N$, we necessarily have:
$p(b | X) =1/\big|[B]^N\big|=B^{-N}$.

\subsection{Derivation of {\boldmath $U(\theta)$} and {\boldmath $\nabla U(\theta)$}}
\label{appdx:form-U}

Recall from~(\ref{eq:U}) in Section~\ref{s:sfb} that,
$$	\pi_\theta(\theta) \propto p(\theta | X, b) \propto p(b | X, \theta) p(\theta) \propto  \exp \left( - U(\theta) \right),
$$ 
so that $U$ can be expressed as,
$$
U(\theta) 
= - \left( \log p(b | X, \theta) + \log p(\theta) \right) + \textrm{const}.
$$
Writing,
$y_{ij} \coloneqq \one \left\{ b_i = j \right\}$ and 
$a_{ij} \coloneqq \phi_j(x_i; \theta)$ as before, we have that,
\begin{equation*}
	\log p(b | X, \theta) = \sum_{i \in [N]} \sum_{j \in [B]} y_{ij} \log a_{ij}  \quad \textrm{and} \quad
	\log p(\theta) = -\frac{D B}{2} \log 2\pi - \frac{1}{2 \sigma_\theta^2} 
	\|\theta \|^2,
	\label{eqn:U-constituent-terms}
\end{equation*}
where
$\|\theta\|^2 = \sum_{i} \theta_{i}^2 = \sum_{j \in [B]} \|w_j\|^2$ 
is the Euclidean norm of the vector of parameters $\theta$.
Therefore, discarding constant terms, we 
obtain,
\begin{equation}
	U(\theta) = \left( \sum_{i \in [N]} \sum_{j \in [B]} y_{ij} \log \frac{1}{a_{ij}} \right)
	+ \frac{1}{2\sigma_\theta^2} \|\theta\|^2.
	\label{eqn:U-form-appdx}
\end{equation}
Now to find $\nabla U(\theta)$, we need to compute
each of its components,
$\nicefrac{\partial U}{\partial w_k}$, for $k \in [B]$.
To that end, we first compute,
\begin{align}
	\frac{\partial a_{ij}}{\partial w_k} &= \frac
	{x_i \exp(w_j^T x_i) \delta_{jk} \cdot \sum_{r \in [B]} \exp(w_r^T x_i) 
		- 
		\exp(w_j^T x_i) \cdot x_i \exp(w_k^T x_i)}
	{\left( \sum_{r \in [B]} \exp(w_r^T x_i) \right)^2} \nonumber \\
	&= x_i \left( a_{ij} \delta_{jk} - a_{ij}a_{ik} \right), 
	\label{eq:dadw}
\end{align}
where $\delta_{jk} \coloneqq \one \left\{ j = k \right\}$,
and we also easily find,
\begin{equation}
	\frac{ \partial}{\partial w_k} \|\theta\|^2 = \frac{\partial}{\partial w_k} \left( \sum_{r \in [B]} \|w_r\|^2 \right) = 2w_k.
	\label{eq:dtsdw}
\end{equation}
Combining the expression for $U(\theta)$ in~(\ref{eqn:U-form-appdx}) and the expressions of~(\ref{eq:dadw}) and~(\ref{eq:dtsdw}), we obtain,
\begin{align}
	\frac{\partial U}{\partial w_k} &= 
	\sum_{i \in [N]} \sum_{j \in [B]} y_{ij} 
	\left( -\frac{x_i}{a_{ij}} \left( a_{ij} \delta_{jk} - a_{ij} a_{ik} \right) \right)
	+ \frac{w_k}{\sigma_\theta^2} \nonumber \\
	&= - \left( \sum_{i \in [N]} \Big\{ x_i (y_{ik} - a_{ik}) \Big\} - \frac{w_k}{\sigma_\theta^2} \right).
\end{align}
This can be computed 
efficiently through matrix operations. The only property of $y_{ij}$ 
we have used in the derivation is the constraint $\sum_{j \in [B]} y_{ij} = 1$,
for all $i$.

\subsection{Implementation details}
\label{appdx:hyperparams}

Full source code is available at:\\
\centerline{\url{https://github.com/LozzaTray/Jormungandr-code}}
Table~\ref{tab:hyperparams} contains all 
hyper-parameter values used in our experiments. The set of retained 
samples are generated as,
$
	\Tcal_\star = \{T_\star \kappa_\star + i \lambda_\star :  
	0 \leq i \leq \lfloor T_\star(1 - \kappa_\star) / \lambda_\star \rfloor \}.
$
\begin{table}[!ht]
	\centering
	\caption{Hyper-parameter values for each experiment.}
	\label{tab:hyperparams}
	\begin{tabular}{c|ccc|ccc|ccc|cc|ccc}
		Dataset & 
		$B$ & $f$ & $\sigma_\theta$ & 
		$T_b$ & $\kappa_b$ & $\lambda_b$ & 
		$T_\theta$ & $\kappa_\theta$ & $\lambda_\theta$ & 
		$k$ & $D'$ &
		$T_\theta'$ & $\kappa_\theta'$ & $\lambda_\theta'$ 
		\\ \hline
		Polbooks &
		3 & 0.7 & 1 &
		1,000 & 0.2 & 5 &
		10,000 & 0.4 & 10 & 
		-- & -- & 
		-- & -- & -- 
		\\
		School &
		10 & 0.7 & 1 &
		1,000 & 0.2 & 5 &
		10,000 & 0.4 & 10 & 
		1 & 10 & 
		10,000 & 0.4 & 10 
		\\
		FB egonet &
		10 & 0.7 & 1 &
		1,000 & 0.2 & 5 &
		10,000 & 0.4 & 10 & 
		1 & 10 & 
		10,000 & 0.4 & 10 
		\\
	\end{tabular}
\end{table}

	\bibliography{sources.bib}

\begin{thebibliography}{10}
\providecommand{\url}[1]{\texttt{#1}}
\providecommand{\urlprefix}{URL }
\providecommand{\doi}[1]{https://doi.org/#1}

\bibitem{cluster-compression}
Abbe, E.: Graph compression: The effect of clusters. In: 54th Annual Allerton
  Conference on Communication, Control, and Computing. pp.~1--8 (2016)

\bibitem{mixed-membership-sbm}
Airoldi, E.M., Blei, D., Fienberg, S., Xing, E.: Mixed membership stochastic
  blockmodels. In: Koller, D., Schuurmans, D., Bengio, Y., Bottou, L. (eds.)
  Advances in Neural Information Processing Systems. vol.~21 (2009)

\bibitem{pseudo-marginal}
Andrieu, C., Roberts, G.O.: The pseudo-marginal approach for efficient {Monte
  Carlo} computations. The Annals of Statistics  \textbf{37}(2),  697--725
  (2009)

\bibitem{link-prediction}
Gaucher, S., Klopp, O., Robin, G.: Outliers detection in networks with missing
  links. Computational Statistics \& Data Analysis  \textbf{164},  107308
  (2021)

\bibitem{hastings-alg}
Hastings, W.K.: {Monte Carlo} sampling methods using {Markov} chains and their
  applications. Biometrika  \textbf{57}(1),  97--109 (1970)

\bibitem{fb-snap}
Leskovec, J., Mcauley, J.: Learning to discover social circles in ego networks.
  In: Pereira, F., Burges, C.J.C., Bottou, L., Weinberger, K.Q. (eds.) Advances
  in Neural Information Processing Systems. vol.~25 (2012)

\bibitem{vanilla-sbm}
Nowicki, K., Snijders, T.A.B.: Estimation and prediction for stochastic
  blockstructures. Journal of the American Statistical Association
  \textbf{96}(455),  1077--1087 (2001)

\bibitem{polbooks}
Pasternak, B., Ivask, I.: Four unpublished letters. Books Abroad
  \textbf{44}(2),  196--200 (1970)

\bibitem{Peixoto-MCMC}
Peixoto, T.P.: Efficient {Monte Carlo} and greedy heuristic for the inference
  of stochastic block models. Physical Review E  \textbf{89}(1) (2014)

\bibitem{Peixoto-Bayesian-Microcanonical}
Peixoto, T.P.: Nonparametric {Bayesian} inference of the microcanonical
  stochastic block model. Physical Review E  \textbf{95}(1) (2017)

\bibitem{peixoto_graph-tool_2014}
Peixoto, T.: The {\tt graph-tool} {Python} library. figshare  (2014),
  \url{figshare.com/articles/graph_tool/1164194}

\bibitem{mala-tweedie}
Roberts, G.O., Tweedie, R.L.: {Exponential convergence of Langevin
  distributions and their discrete approximations}. Bernoulli  \textbf{2}(4),
  341--363 (1996)

\bibitem{schools}
Stehlé, J., Voirin, N., Barrat, A., Cattuto, C., Isella, L., Pinton, J.F.,
  Quaggiotto, M., Van~den Broeck, W., Régis, C., Lina, B., Vanhems, P.:
  High-resolution measurements of face-to-face contact patterns in a primary
  school. PLoS ONE  \textbf{6}(8),  1--13 (2011)

\bibitem{overlapping-sbm}
Zhu, J., Song, J., Chen, B.: Max-margin nonparametric latent feature models for
  link prediction. arXiv preprint {\tt cs.LG:1602.07428} (2016)

\end{thebibliography}
		
\end{document}